\documentclass[conference]{IEEEtran}
\IEEEoverridecommandlockouts
\usepackage{cite}
\usepackage{amsmath,amssymb,amsfonts}
\usepackage{algorithmic}
\usepackage{graphicx}
\usepackage{textcomp}
\usepackage{booktabs}
\usepackage{xcolor}
\def\BibTeX{{\rm B\kern-.05em{\sc i\kern-.025em b}\kern-.08em
    T\kern-.1667em\lower.7ex\hbox{E}\kern-.125emX}}
\usepackage{subcaption}
\usepackage[hidelinks]{hyperref}

\begin{document}

\title{Analysis of the Effect of Low-Overhead\\Lossy Image Compression on the Performance of Visual Crowd Counting for Smart City Applications\\
\thanks{This work was funded by the European Union’s Horizon 2020 research and innovation programme under grant agreement No 957337. This publication reflects the authors views only. The European Commission is not responsible for any use that may be made of the information it contains.}
}

\author{\IEEEauthorblockN{Arian Bakhtiarnia, Błażej Leporowski, Lukas Esterle and Alexandros Iosifidis}
\IEEEauthorblockA{\textit{Department of Electrical and Computer Engineering, Aarhus University, Denmark}}
}

\maketitle

\begin{abstract}
Images and video frames captured by cameras placed throughout smart cities are often transmitted over the network to a server to be processed by deep neural networks for various tasks. Transmission of raw images, i.e., without any form of compression, requires high bandwidth and can lead to congestion issues and delays in transmission. The use of lossy image compression techniques can reduce the quality of the images, leading to accuracy degradation. In this paper, we analyze the effect of applying low-overhead lossy image compression methods on the accuracy of visual crowd counting, and measure the trade-off between bandwidth reduction and the obtained accuracy.
\end{abstract}

\section{Introduction}

Many tasks in smart cities, such as visual crowd counting, rely on processing images and videos captured by surveillance cameras placed throughout the city. Typically, the input captured by several cameras is transmitted to an edge server or a cloud server in order to be processed by deep neural networks (DNNs). Since modern cameras can capture visual information in high resolutions, in cases even exceeding Full HD ($ 1920 \times 1080 $ pixels), transmitting raw video frames of the (real-time) visual stream over the network requires massive amounts of bandwidth and leads to congestion issues. %To reduce bandwidth consumption, the size of raw input video frames can be reduced by decreasing their quality. However, depending on the visual data analysis task at hand, this can have a negative impact on the accuracy. 
When using DNNs for visual data analysis, the adoption of a specific neural network architecture can guide decisions related to the size of images to be processed as some DNNs require inputs of a fixed size. For instance, the input size of image classification DNNs is typically fixed to $ 224 \times 224 $ or $ 384 \times 384 $ pixels. In these cases, preserving the sensor image resolution during transmission is not sensible, since the image will be resized to the DNN input size regardless. Therefore, downsizing the image to the target resolution before transmission is a simple and low-overhead solution that will not negatively impact the accuracy. 

However, depending on the visual data analysis task at hand, the above described process can have a negative impact on the accuracy. Dense classification tasks which output a classification result for each input pixel, such as semantic image segmentation, and dense regression tasks which output a heatmap that assigns a number to each input pixel, such as visual crowd counting, do not necessarily require a fixed input size. More generally, when the adopted DNN architecture has the properties of fully-convolutional neural networks (FCNs) in which the size of the output can change depending on the input size, reducing the size and quality of inputs to these DNNs can have a drastic impact on their accuracy.

Reduction of the input image size can, in principle, be obtained by applying any lossy image compression method. However, different methods will lead to different types of visual quality reduction for lowering the amount of bandwidth needed to transmit the compressed video frames. Thus, they can have a different impact on the accuracy achieved by the DNN used to solve the visual analysis task\footnote{Hereafter, we refer to the DNN used to provide a solution to the visual analysis task as the task DNN.}. In order to achieve a good compromise between high image compression and high accuracy on the visual analysis task at hand, the effect of using different lossy image compression methods needs to be carefully investigated. Such an investigation needs to also consider the computational cost required for applying the different compression methods. This is due to the fact that compression needs to be applied before transmission, and cameras (or IoT devices attached to the cameras) may not possess the capability to perform computationally expensive compression operations.

In this paper, we provide an analysis of the effects that lossy image compression can have on the performance of visual crowd counting based on DNNs. We assume that the adopted IoT devices have very limited computational resources and are therefore only capable of performing low-overhead compression operations. We investigate three low-computational compression operations, namely uniform downsampling, JPEG compression, and grayscaling, we recount the advantages and disadvantages of each approach and measure the trade-off between accuracy and bandwidth reduction for each method. We show that for the task of crowd counting, JPEG compression leads to the best trade-off.

The rest of this paper is organized as follows. Section \ref{sec:related_work} describes related works in visual crowd counting, as well as lossy image compression. Section \ref{sec:methods} provides details on the three compression methods investigated in this work. Section \ref{sec:experiments} provides the experimental setup and results of our experiments. Finally, section \ref{sec:conclusion} concludes the paper by discussing the experimental results, and provides some future directions. Our code is available at \url{https://github.com/CptPirx/vcc-compression}.

\section{Related Work}\label{sec:related_work}

\subsection{Visual crowd Counting}
\begin{figure}
\centering
\begin{subfigure}{.24\textwidth}
  \centering
  \includegraphics[width=.98\linewidth]{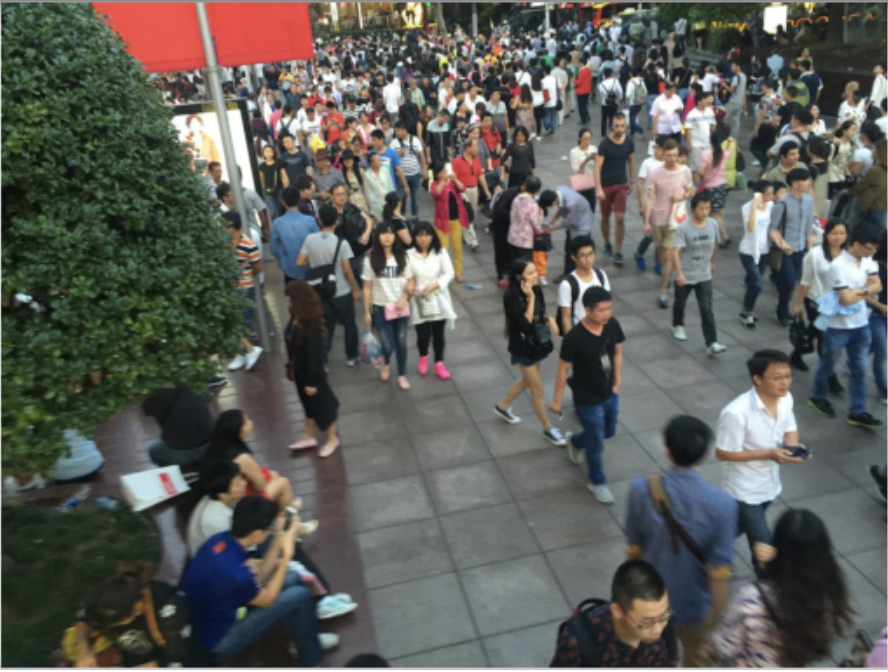}
\end{subfigure}%
\begin{subfigure}{.24\textwidth}
  \centering
  \includegraphics[width=.98\linewidth]{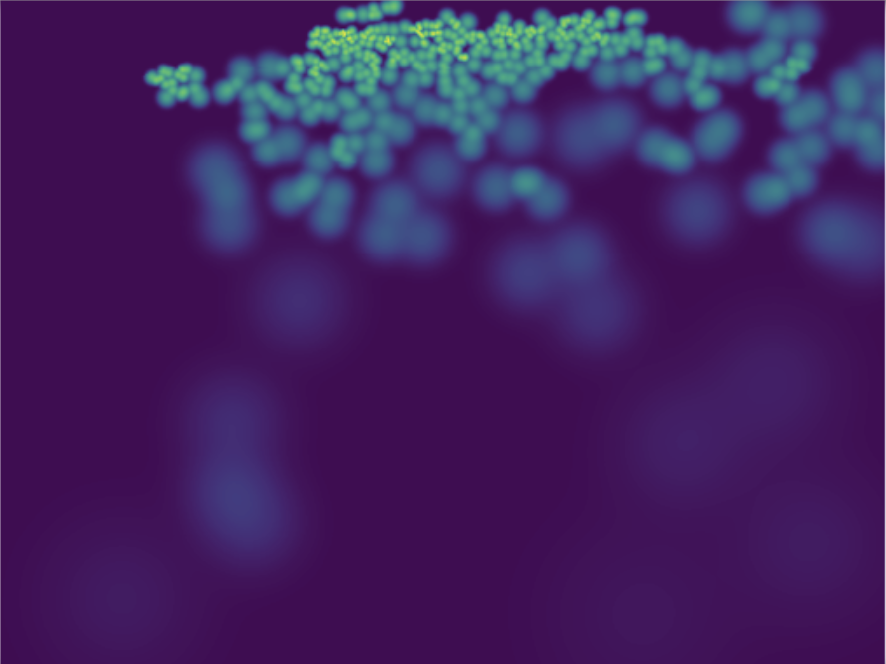}
\end{subfigure}
\caption{An example image from the Shanghai Tech dataset (left) \cite{zhang2016single} and its corresponding ground truth density map (right).}
\label{fig:crowd_counting_label_example}
\end{figure}

Visual crowd counting is the task of counting the total number of people present in a scene given an image of that scene. In the context of smart cities, visual crowd counting can be used to monitor crowds, collect crowd statistics over time, and increase safety in areas of particular interest \cite{bajovic2021marvel}. The input to a visual crowd counting method is an image or video frame of the scene, and the output is a single number representing the total count. However, most crowd counting methods also provide a density map as output which specifies the density of people at each location of the input, as shown in figure \ref{fig:crowd_counting_label_example}. The annotations available in most visual crowd counting datasets are in the form of head annotations, where the location of the center of the head of each person in the scene is specified. Variations in illumination and perspective, as well as partial occlusions can make crowd counting a challenging task.

Various DNN-based approaches for crowd counting exist in the literature \cite{gao2020cnn}. In this work, we use SASNet \cite{song2021choose}, which at the time of writing is the state-of-the-art method for visual crowd counting on the Shanghai Tech dataset \cite{zhang2016single}, a very popular dataset for evaluating the performance of crowd counting methods.

\subsection{Lossy image compression for DNNs}
Many lossy image compression methods exist that can be used to decrease the size of inputs to DNNs. However, most of these methods are computationally expensive and add a lot of overhead to the task. Furthermore, they require great effort in design and training for the specific setting. For instance, non-uniform downsampling (NUD) methods downsize the image to lower resolutions while sampling more pixels from salient areas of the image, leading to a distorted compressed image \cite{recasens2018learning}. Therefore, NUD requires saliency detection as a first step, which is typically performed using another DNN and is, thus, computationally expensive. Moreover, DNN used to perform saliency detection needs to be trained for detecting visual saliency related to the specific task at hand, and have a relatively high accuracy in order for NUD to be effective.

Another example would be neural image compression (NIC) methods encoding input images to compact representations which can be transmitted instead of the raw input images \cite{tellez2019neural}. However, the encoder in NIC is also a DNN, leading to the same challenges described above for the NUD methods related to high computational cost and the need of extensive training. Furthermore, multilinear compressive learning compresses the input tensor before transmission to the server side \cite{9070152,tran2022IoT}. However, this method has been shown to perform well in cases where the input resolution is low or the inputs are very similar in nature and have specific characteristics, for instance, they are all facial images. Therefore, this method may not be useful in the smart cities setting discussed in this paper.

The methods explored in this work and described in section \ref{sec:methods} are simple solutions that have a low added computational overhead, and can be easily plugged into any existing edge intelligence system. An additional benefit of these methods is that they are readily available as configurations in most cameras, and may not need additional computational resources, such as Raspberry Pi, to process the frames after they are captured.

\section{Low-Overhead Image Compression Methods for Efficient Crowd Counting}\label{sec:methods}

\subsection{Uniform Downsampling}
Uniform downsampling performs a simple resizing of the image to a lower resolution, where sampling is performed uniformly across the image. An advantage of this method is that the amount of compression, and thus the trade-off between bandwidth reduction and image quality degradation, can be tuned by specifying the target resolution. However, uniform downsampling can lead to a significant drop in accuracy in visual crowd counting when the resolution is significantly reduced. This is due to that the heads of multiple people in the scene may be represented by a single pixel, making it impossible to accurately determine the number of people in that location. Note that when the aspect ratio of the target resolution, that is, the proportional relationship between the width and height of the image, is different from the aspect ratio of the input image, uniform downsampling can lead to harmful image distortions. For instance, in crowd counting, the shape of people's heads can be elongated, which can make it difficult for the DNN to detect them.

Several approaches exist for uniformly sampling pixels from a high-resolution image into a lower-resolution one. \textit{Nearest} simply takes the nearest corresponding pixel from the high-resolution image, and ignores all other pixels. \textit{Bilinear} and \textit{Bicubic} use linear and cubic interpolation, respectively, on all pixels from the high-resolution that may contribute to the output pixel value. Finally, \textit{Lanczos} calculates the output pixel value using a high-quality Lanczos filter \cite{turkowski1990filters}.

\subsection{JPEG Compression}\label{sec:jpeg}
JPEG encoding consists of three steps \cite{7924246, gueguen2018faster}. The first step takes the 3-channel 24-bit RGB image as input and converts it to the YCbCr color space based on the following formula
\begin{small}
\begin{multline}
\begin{bmatrix}
Y\\
Cb\\
Cr
\end{bmatrix}
=
\begin{bmatrix}
0.299 & 0.587 & 0.114\\
-0.168935 & -0.331665 & 0.50059\\
0.499813 & -0.418531 & -0.081282
\end{bmatrix}
\begin{bmatrix}
R\\
G\\
B
\end{bmatrix}.
\label{eq:rgb2ycbcr}
\end{multline}
\end{small}

\begin{figure}
\begin{center}
\begin{tabular}{ c c }
\includegraphics[width=.23\textwidth]{images/cc_img.png}&
\includegraphics[width=.23\textwidth]{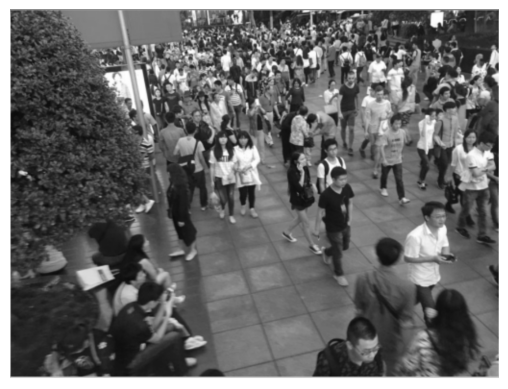}\\
(a)&(b)\\
\includegraphics[width=.23\textwidth]{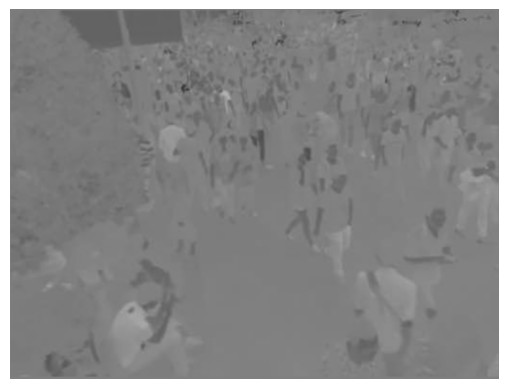}&
\includegraphics[width=.23\textwidth]{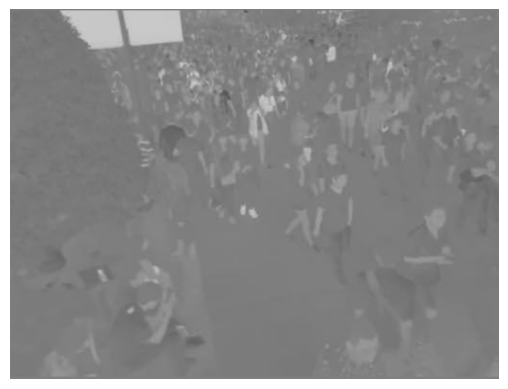}\\
(c)&(d)
\end{tabular}
\end{center}
\caption{(a) Original image; (b) Y component; (c) Cb component; and (d) Cr component.}
\label{fig:ycbcr}
\end{figure}

\noindent The Y component, called \textit{luma}, represents the brightness, and the Cb and Cr components called \textit{chroma} represent color. Since the human eye is less sensitive to fine color detail, the resolution of chroma components is reduced by a factor of 2 or 3. Figure \ref{fig:ycbcr} shows an example image and its corresponding Y, Cb and Cr components. In the second step, each component is split into blocks of size $ 8 \times 8 $ which undergo a two-dimensional \textit{discrete cosine transform (DCT)}. DCT converts the spatial domain into frequency domain. The amplitude of the frequency domain is then quantized based on
\begin{equation}
Q = D \oslash T_s,
\label{eq:quantize_hadamard}
\end{equation}
where $ D $ is the non-quantized DCT matrix, $ Q $ is the quantized result, $ \oslash $ denotes Hadamard division (matrix element-wise division), and $ T_s $ is derived based on
\begin{equation}
T_{s_{ij}} = \Big\lfloor \frac{s T_{b_{ij}} + 50}{100} \Big\rfloor,
\label{eq:quantize_ts}
\end{equation}
where
\begin{equation}
s = \begin{cases}
  \frac{5000}{q}, & 1 \leq q < 50, \\
  200 - 2q, & 50 \leq q \leq 100,
\end{cases}
\label{eq:quantize_s}
\end{equation}
and $ T_b $ is the base quantization matrix
\begin{equation}
T_b = \begin{bmatrix}
16 & 11 & 10 & 16 & 24 & 40 & 51 & 61\\
12 & 12 & 14 & 19 & 26 & 58 & 60 & 55\\
14 & 13 & 16 & 24 & 40 & 57 & 69 & 56\\
14 & 17 & 22 & 29 & 51 & 87 & 80 & 62\\
18 & 22 & 37 & 56 & 68 & 109 & 103 & 77\\
24 & 35 & 55 & 64 & 81 & 104 & 113 & 92\\
49 & 64 & 78 & 87 & 103 & 121 & 120 & 101\\
72 & 92 & 95 & 98 & 112 & 100 & 103 & 99
\end{bmatrix}.
\label{eq:quantize_tb}
\end{equation}
All zeroes in $ T_s $ are converted to one to prevent division by zero. The quality setting $ 1 <= q <= 100 $ in equation \ref{eq:quantize_s} is an integer controlling the quality of the compression, where 1 is the lowest quality and 100 the highest. Finally, the size of quantized matrix is further reduced using Huffman encoding, which is a lossless compression algorithm.

On the server side, the JPEG is decoded back to RGB by reversing these steps and processed using the task DNN. Even though neural networks that directly use JPEG features instead of RGB exist in the literature and are slightly more efficient than converting JPEG back to RGB \cite{gueguen2018faster}, designing such neural networks is not a simple task \cite{goldberg2020rethinking}.

An advantage of using JPEG compression is that the quality setting can be tuned to control the trade-off between bandwidth and visual quality. However, a drawback of this approach compared to the two other approaches is that exact amount of compression depends on each specific image and can vary for different input images.

\subsection{Grayscaling}
The number of channels in the image can be reduced from three down to one by converting the color image into a grayscale image. The significance of color information varies across deep learning tasks. However, the fact grayscale data augmentation is widely used in computer vision task, shows that a lot of information can still be obtained from colorless images. The luminance component (Y) in YCbCr color space described in section \ref{sec:jpeg} is essentially a grayscale version of the image. Therefore, conversion to grayscale can be obtained by dropping the Cb and Cr channels, as shown in figure \ref{fig:ycbcr} (b). We use the standard grayscale transform in PyTorch \cite{paszke2019pytorch}, which follows the same approach, but with YPbPr color space instead, which is the gamma-corrected version of YCbCr. The main disadvantage of the grayscaling method is that, unlike previous approaches, the amount of compression is not tunable and the bandwidth is always decreased by a factor of three.

Note that even though the task DNN on the server receives RGB images with three channels as input, the architecture of this DNN does not need to change, since the information in the grayscale channel can be copied to all three channels. This trick has been shown to work for processing single-channel audio spectrograms with DNNs designed for processing RGB images \cite{BAKHTIARNIA2022461}.

\section{Experiments}\label{sec:experiments}
We conducted experiments on the Shanghai Tech dataset \cite{zhang2016single}, which is a widely adopted dataset for evaluating the performance of visual crowd counting methods. 
It contains images of resolution $1024\times768$ pixels which are already compressed with JPEG at quality $q$ of $75$.  

For our experiments we treat the memory size of the original image, stored in a numpy integer array.
In order to evaluate the effect of different compression rates in the accuracy of visual crowd counting, we applied uniform downsampling and JPEG compression methods in multiple steps.
For JPEG compression, at each step the quality setting $q$ has been decreased by $5$. 
Starting from the value of $q=75$, this being the compression rate of the original images in the dataset, down to the value of $q=5$. 
%As the original images in the dataset are already compressed with JPEG, which is a lossy compression method, it is impossible to achieve a better quality image than the original source just by using different quality setting.
For uniform downsampling, at each step the image dimensions where reduced in by $10\%$ of the original image size. 
Starting from $100\%$, i.e., the dimensions of $1024\times768$ pixels of the original images in the dataset, down to $102\times76$ pixels, i.e., $10\%$ of the original image dimensions. 
The effect of using the grayscale version of the original images to the accuracy of visual crowd counting is measured once, i.e., by transforming the original images in the dataset to their grayscale version. 

The process followed to finetune the SasNet DNN \cite{song2021choose} on each experiment follows the same protocol. 
We used a Linux server with four RTX 2080 Ti GPUs. On each experiment, the compressed images are obtained and the pretrained weights of the DNN model trained on the Shanghai Tech dataset are finetuned for $100$ epochs using the compressed images with a mini-batch size of $4$ using the AdamW optimizer \cite{DBLP:conf/iclr/LoshchilovH19} with a learning rate of $ 10^{-5} $ and a weight decay of $ 10^{-4} $.

\begin{figure}
    \centering
    \includegraphics[width=0.5\textwidth]{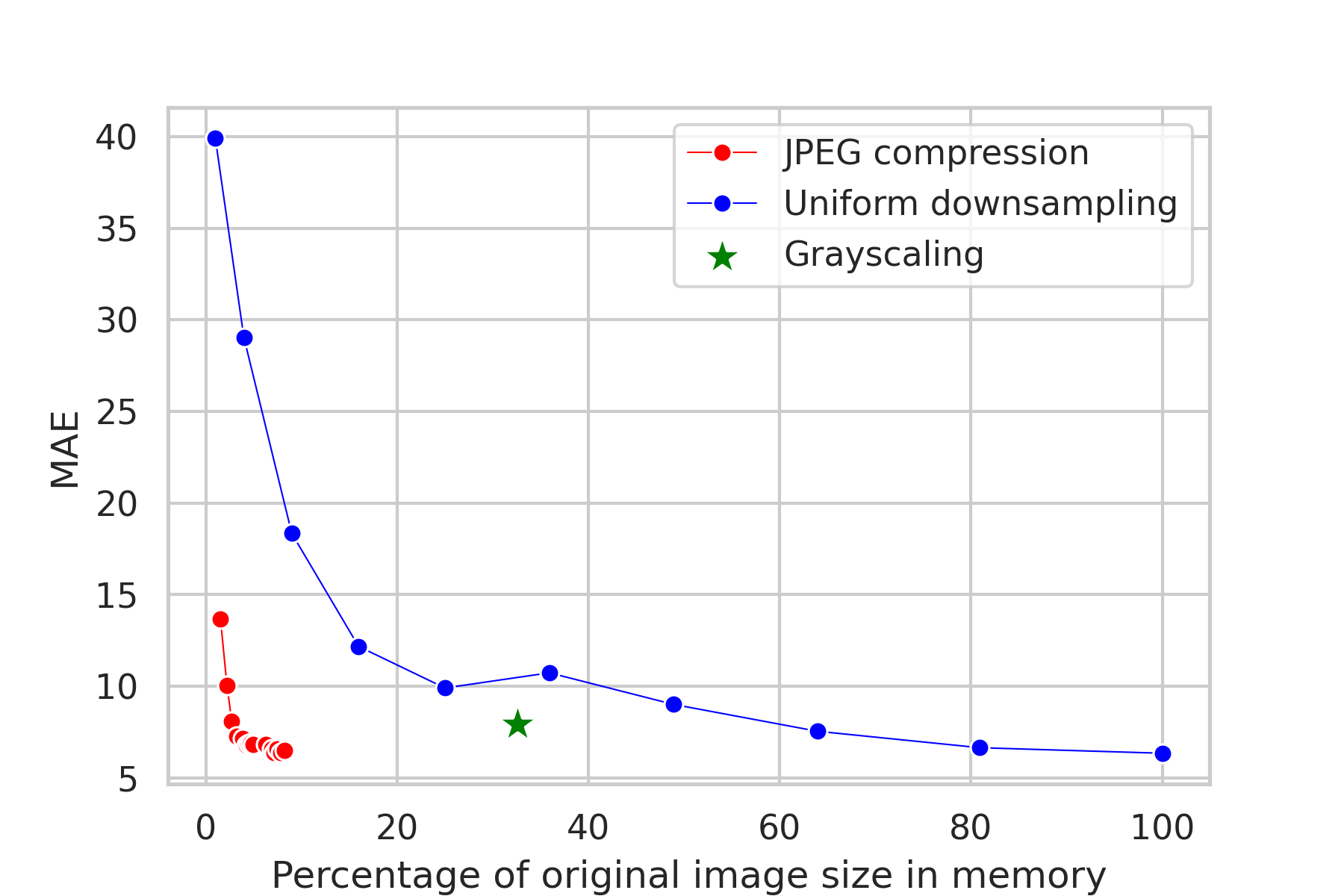}
    \caption{Visual crowd counting performance (MAE) with different data size reduction methods}
    \label{fig:mae_plot}
\end{figure}

\begin{figure}
    \centering
    \includegraphics[width=0.5\textwidth]{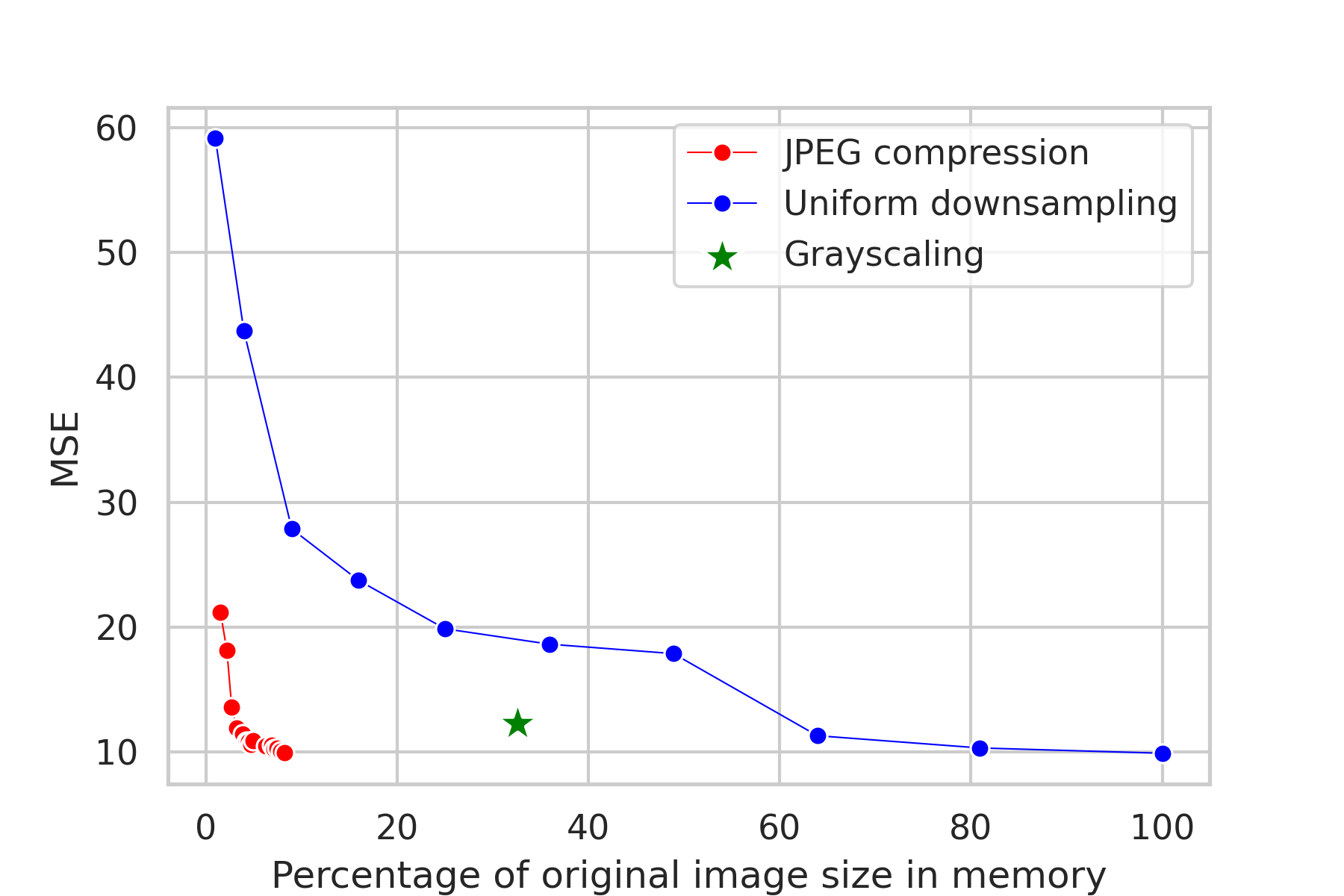}
    \caption{Visual crowd counting performance (MSE) with different data size reduction methods}
    \label{fig:mse_plot}
\end{figure}

The test results are reported for the best performing model obtained during those 100 epochs, which may not necessarily be the model after 100 epochs. Figures \ref{fig:mae_plot} and \ref{fig:mse_plot} show the performance obtained when using the mean absolute error (MAE) and the mean squared error (MSE) metrics, respectively. The results are consistent across both MAE and MSE metrics. The horizontal axis represents the in-memory size of the image in relation to the size of the original image.

\begin{table}[h]
    \centering
    \caption{Visual crowd counting performance using JPEG compression.}
    \begin{tabular}{cccc}
        \toprule
        {Quality} & MAE & MSE & Size \\
        \midrule
        75  & 6.35  & 9.90  & 8.16\%  \\
        70  & 6.37  & 10.04 & 7.83\%  \\
        65  & 6.58  & 10.29 & 7.40\%  \\
        60  & 6.40  & 10.23 & 7.13\%  \\
        55  & 6.57  & 10.56 & 6.91\%  \\
        50  & 6.83  & 10.50 & 6.28\%  \\
        45  & 6.83  & 10.91 & 4.92\%  \\
        40  & 6.91  & 10.61 & 4.68\%  \\
        35  & 6.86  & 10.83 & 4.44\%  \\
        30  & 6.80  & 10.83 & 4.26\%  \\
        25  & 7.14  & 11.43 & 3.86\%  \\
        20  & 7.29  & 11.93 & 3.24\%  \\
        15  & 8.10  & 13.57 & 2.70\%  \\
        10  & 10.07 & 18.11 & 2.20\%  \\
        5   & 13.67 & 21.17 & 1.52\%  \\
        \bottomrule
    \end{tabular}
\label{tab:compression_results}
\end{table}

\begin{table}[h]
    \centering
    \caption{Visual crowd counting performance using uniform downsampling.}
    \begin{tabular}{cccc}
        \toprule
        {Dimensions as \% of original} & MAE & MSE & Size \\
        \midrule
        100 & 6.35  & 9.90  & 100\%  \\
        90  & 6.65  & 10.32 & 80.92\%  \\
        80  & 7.55  & 10.30 & 63.94\%  \\
        70  & 9.02  & 17.88 & 48.89\%  \\
        60  & 10.74 & 18.63 & 35.91\%  \\
        50  & 9.91  & 19.87 & 25\%  \\
        40  & 12.18 & 23.74 & 15.97\%  \\
        30  & 18.34 & 27.86 & 8.98\%  \\
        20  & 29.04 & 43.72 & 3.97\%  \\
        10  & 39.90 & 59.14 & 0.99\%  \\
        \bottomrule
    \end{tabular}
\label{tab:downsampling_results}
\end{table}

Our experiments show that decreasing resolution has a much higher impact on the performance of the visual crowd counting model than decreasing image quality or removing the color components. 
The drop in performance when decreasing the image resolution is small, down to the value of $80\%$ of the original dimensions, i.e., $819\times614$ pixels.
Decreasing the image resolution more leads to higher performance losses. 
When the image dimensions are decreased below $40\%$ of the original image size, i.e., $409\times307$ pixels, the performance starts to decrease significantly. 
An image with these dimensions has a memory size of $15.97\%$ of the original image. 
The decrease in the performance of visual crowd counting when JPEG compression is used is minimal up to the value of $q=30$, which corresponds to an in-memory image size of $~4.2\%$ of the original image size. 
Below that value the performance starts to decrease in a more rapid manner, but it is still significantly lower compared to applying uniform downsampling.  
Grayscaling results in a small performance drop. 
However, it leads to lower performance compared to JPEG compression, while having a much higher in-memory image size.

The complete results obtained by applying JPEG compression and uniform donwsampling are presented in Tables \ref{tab:compression_results} and \ref{tab:downsampling_results}. 
In all tables the size is given as the percentage of the original image size.
Grayscaling resulted to an MAE value of $7.92$, an MSE value of $12.25$, and an image size of $23.57\%$ of the original image.

Furthermore, we conducted an experiment to determine whether the resampling filter used in the uniform downsampling method impacts the performance of the model. 
The results can be seen in Table \ref{tab:alg_impact}.

\begin{table}[h]
    \centering
    \caption{Impact on visual crowd counting performance of the resampling technique used in uniform downsampling.}
    \begin{tabular}{l|ccc}
        \toprule
        {Dimensions as \% of original} & Method & MAE & MSE \\
        \midrule
        60  & Bicubic   & 10.88 & 19.54  \\
            & Bilinear  & 10.34 & 18.82  \\
            & Lanczos   & 10.45 & 18.10  \\
            & Nearest   & 10.74 & 18.63  \\ \hline
        30  & Bicubic   & 19.11 & 30.37  \\
            & Bilinear  & 16.62 & 26.73  \\
            & Lanczos   & 18.87 & 30.12  \\
            & Nearest   & 18.34 & 27.86  \\
        \bottomrule
    \end{tabular}
\label{tab:alg_impact}
\end{table}
At 60\% of the original image dimensions, the resumbling methods lead to similar visual crowd counting performance. The biggest difference in performance when considering the MAE metric is between Bicubic and Bilinear methods. When considering the MSE metric, the biggest performance difference is between Bicubic and Lanczos methods. 
At 30\% of the original image dimensions the differences in the visual crowd counting performance are more substantial. 
%For MAE, the biggest difference is again between the Bicubic and Bilinear methods, $13\%$.
%The same pair of methods has the biggest difference in MSE metric, $12\%$.

\section{Conclusion}\label{sec:conclusion}
In this paper, we provided an analysis of the effect of applying three low-overhead lossy image compression methods to the performance of visual crowd counting in the case where the capture devices have limited computational resources. Based on experiments conducted in a widely used public dataset, we observed that JPEG compression provides the best trade-off between accuracy and bandwidth reduction. This implies that even though the most intuitive and widely used approach in deep learning is uniform downsampling, when faced with bandwidth limitations, practitioners are better off compressing their images and video frames with JPEG. As seen in figure \ref{fig:mae_plot}, even grayscaling provides a better trade-off between accuracy and bandwidth reduction compared to uniform downsampling, albeit with only one option to choose from. While the advantages of applying JPEG over uniform downsampling or grayscaling are clear in visual crowd counting task, it should be noted that different visual analysis tasks can have different properties and thus, the the same conclusions may not be drawn for other tasks.

As future directions for this research, other low-overhead lossy image compression methods, such as those based on wavelet transform \cite{wang2022image}, could be investigated. JPEG is optimized for providing high visual quality to human observers, however, it may not be the best compression algorithm for deep neural networks performing a particular task. Moreover, combinations of compression methods can be investigated, for instance, using uniform downsampling and then applying JPEG encoding, or vice versa.

\bibliographystyle{IEEEtran}
\bibliography{references.bib}

\end{document}